%% file: main.tex
\documentclass[runningheads]{llncs}

\usepackage{amsmath}
\usepackage{amssymb}
\usepackage{graphicx}
\usepackage{arydshln}
\usepackage{nicefrac}
\usepackage{booktabs}
\usepackage{bm}
\usepackage{pifont}
\usepackage{tikz}
\usepackage{xspace}
\usepackage{wrapfig}
\usepackage{hyperref}
\usepackage{subcaption}

\newcommand{\model}[1]{{\fontfamily{qpl}\selectfont {\small #1}}}

\newcommand{\dataset}[1]{{\fontfamily{ptm}\selectfont {\small #1}}}
\newcommand{\myparagraph}[1]{\vspace{0.5pt}\noindent{\bf #1}}
\newcommand{\xmark}{\ding{55}}

\newcommand{\ie}{i.e.,\ }
\newcommand{\eg}{e.g.,\ }
\newcommand{\wrt}{w.r.t.\ }

\newcommand{\etal}{et al.\@\xspace}

\begin{document}

\title{Privacy Protection in MRI Scans Using 3D Masked Autoencoders}

\author{Lennart Alexander Van der Goten\inst{1,2} \and
Kevin Smith\inst{1,2}}

\authorrunning{Van der Goten \& Smith}

\institute{KTH Royal Institute of Technology, Stockholm, Sweden \and
SciLifeLab, Solna, Sweden}

\maketitle              
\input{0_abstract}
\input{1_introduction}
\input{2_related_work}

\input{3_method}
\input{4_experiments}

\input{5_conclusion}

\clearpage

\bibliographystyle{splncs04}
\bibliography{main}

\end{document}

%% file: 0_abstract.tex
\begin{abstract}
MRI scans provide valuable medical information, however they also contain sensitive and personally identifiable information that needs to be protected. Whereas MRI metadata is easily sanitized, MRI image data is a privacy risk because it contains information to render highly-realistic 3D visualizations of a patient's head, enabling malicious actors to possibly identify the subject by cross-referencing a database.  Data anonymization and de-identification is concerned with ensuring the privacy and confidentiality of individuals' personal information. Traditional MRI de-identification methods remove privacy-sensitive parts (\eg eyes, nose etc.) from a given scan. This comes at the expense of introducing a domain shift that can throw off downstream analyses. 
In this work, we propose \model{CP-MAE}, a model that de-identifies the face by remodeling it (\eg changing the face) rather than by removing parts using masked autoencoders.
\model{CP-MAE} outperforms all previous approaches in terms of downstream task performance as well as de-identification. 
With our method we are able to synthesize high-fidelity scans of resolution up to $256^3$ -- compared to $128^3$ with previous approaches -- which constitutes an eight-fold increase in the number of voxels.

\keywords{Magnetic Resonance Imaging  \and Privacy in Health \and Generative Modeling.}
\end{abstract}

%% file: 1_introduction.tex
\begin{figure}
\begin{center}
\begin{tabular}[t]{@{}c@{}}
\includegraphics[width=0.4\textwidth]{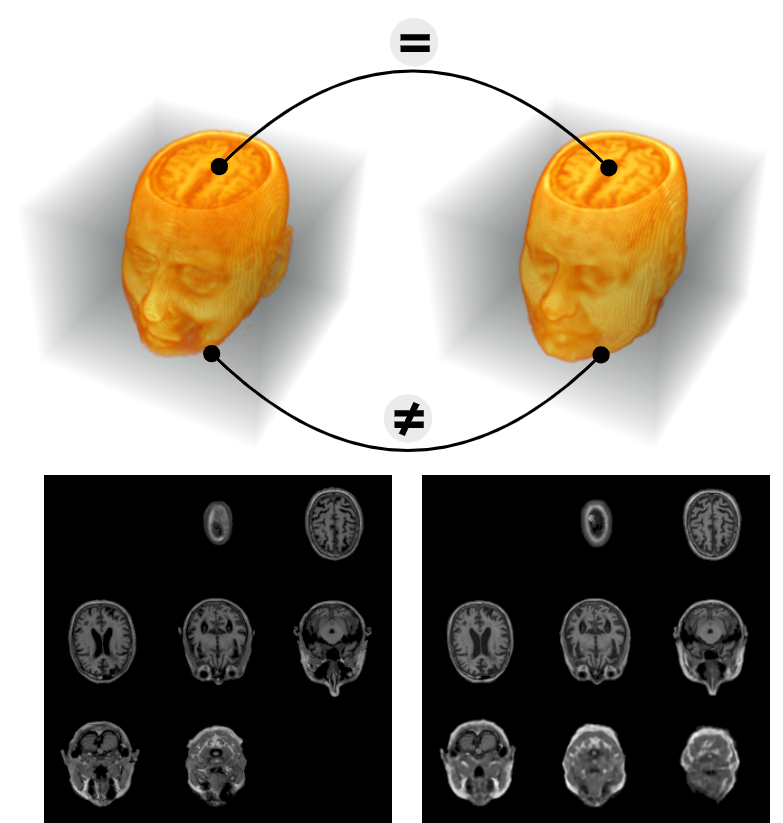} \\
\end{tabular}
\caption{MRI scans pose a privacy risk since highly-realistic face renderings can be crafted and misused for malicious purposes. Our model aims to advance the so-called \emph{remodeling-based} subclass of MRI de-identification which retains the brain ($=$) and remodels all other features ($\neq$). (\emph{top}: 3D view, \emph{bottom} slice-view) }
\label{fig:de-identification}
\end{center}
\end{figure}

\section{Introduction}

While MRI scans are usually visualized as 2D slices, it is also possible to render a high-quality 3D model using techniques such as volumetric raytracing, yielding a realistic depiction of a patient's face. 
This is problematic in terms of data privacy: Given and MRI scan and a face database with associated identities, a malicious actor can find the closest match to a given face rendering, allowing them to \emph{potentially infer} a patient's identity.

Various de-identification methods have appeared over the years that address the removal of privacy-sensitive parts.
Most are quite simple and differ in their level of aggressiveness: While \model{FACE MASK} \cite{milchenko2013facemask} merely blurs out the face, skull-stripping methods such as \model{MRI WATERSHED} \cite{segonne} only retain the brain and remove everything else.
However, these traditional approaches are potentially problematic.
Simply put, many tools in the medical workflow require the presence of certain landmarks as preconditions. 
If those landmarks are absent, the analysis might be impaired or, in the worst case, it might be impossible to perform the analysis at all \cite{de2020facing}. 
On the other hand, if a method is not aggressive enough it might still be feasible to infer a patient's identity, a fact that proves especially harmful when such a tool is to be used automatically and without oversight on a large corpus of scans.

A recently proposed GAN-based approach named \model{CP-GAN} \cite{vandergoten2021conditional} introduced a new paradigm: to remodel the face and leave the medically-relevant information intact.

Simply put, a patient's MRI scan is de-identified by copying their brain and remodeling the remaining parts such that they ``look like'' as if they belong to an actual MRI scan but do not give away information about the patient's real face (or identity). 
This approach eliminates the aforementioned trade-off, as it ensures accurate placement of landmarks and thus minimizes domain shift, leading to a safer and more effective de-identification.

Recently, the emergence of \emph{masked autoencoders} (MAEs) has propelled generative approaches forward, departing from the previously dominant GAN methodology by predicting stochastically-masked segments of the data instead of formulating an adversarial game.

MAEs offer two fundamental advantages over GANs: (i) Higher training stability, and, (ii) Higher data efficiency. Considering that MRI de-identification with generative models involves high memory requirements, data scarcity, and often suffers from instability due to small batch sizes, MAEs appear to be an attractive alternative.

In this work, we adapt masked autoencoders for MRI de-identfication. Our contributions are as follows:
\begin{enumerate}
    \item We propose \model{CP-MAE}, a \emph{masked autoencoder}-based model that can deal with high-resolution MRI scans. Whereas previous remodeling-based method were limited to a resolution of $128^3$, \model{CP-GAN} can produce de-identified scans of size $256^3$, an $8-$fold increase in the number of voxels.
    \item To the best of our knowledge, \model{CP-GAN} is the first to combine a volumetric VQ-VAE with an MAE for 3D MR image synthesis
    \item We demonstrate that \model{CP-MAE} features superior de-identification performance compared to other methods and can be robustly trained on modestly-sized datasets (\eg\  \dataset{ADNI}, \dataset{OASIS-3})
    \item We show that de-identification with \model{CP-MAE} introduces minimal effects on brain tissue and subcortical segmentation tasks compared to other approaches.
\end{enumerate}

%% file: 2_related_work.tex
\section{Related Work}

\myparagraph{MRI De-Identification.}
MRI de-identification is a critical pre-processing step in neuroimaging, and many methods have been proposed to effectively perform this task. 
\model{BET} \cite{smith2002bet} is widely used for its  simplicity. Despite its effectiveness, it sometimes fails to exclude non-brain tissues, such as dura and eyes.
\model{ROBEX} \cite{iglesias2011robex} aims to address some of these shortcomings by employing a trained random forest classifier to distinguish brain and non-brain voxels. 
\model{MRI WATERSHED}, in turn, \cite{segonne} employs a watershed transformation followed by a deformable model to extract the brain.
Milchenko \etal \cite{milchenko2013facemask} propose \model{FACE MASK}, focusing on the protection of sensitive facial information in MRI data. This method combines the benefits of defacing and skull-stripping by generating a mask that can be used to blur out facial features while preserving brain anatomy.
Schimke \etal introduced a tool named \model{QUICKSHEAR} \cite{Schimke2011}, which is intended to compute a hyperplane in 3D space that separates the facial features from the brain. \model{DEFACE} \cite{bischoff2021deface} is a deformable model that estimates which voxels belong to the brain, said voxels can then be cut out for de-identification purposes.

Finally, \model{CP-GAN} \cite{vandergoten2021conditional} remodels non-brain tissues as opposed to removing them. The de-identification process is performed by a conditional GAN that takes as input the patient's brain and a convex hull hinting at the metric extent of the skull to be generated.
Our approach follows this general framework of remodeling the face and skull, as opposed to opting for something that would remove these parts altogether.

\begin{figure*}[t]
    \centering
    \includegraphics[width=\textwidth]{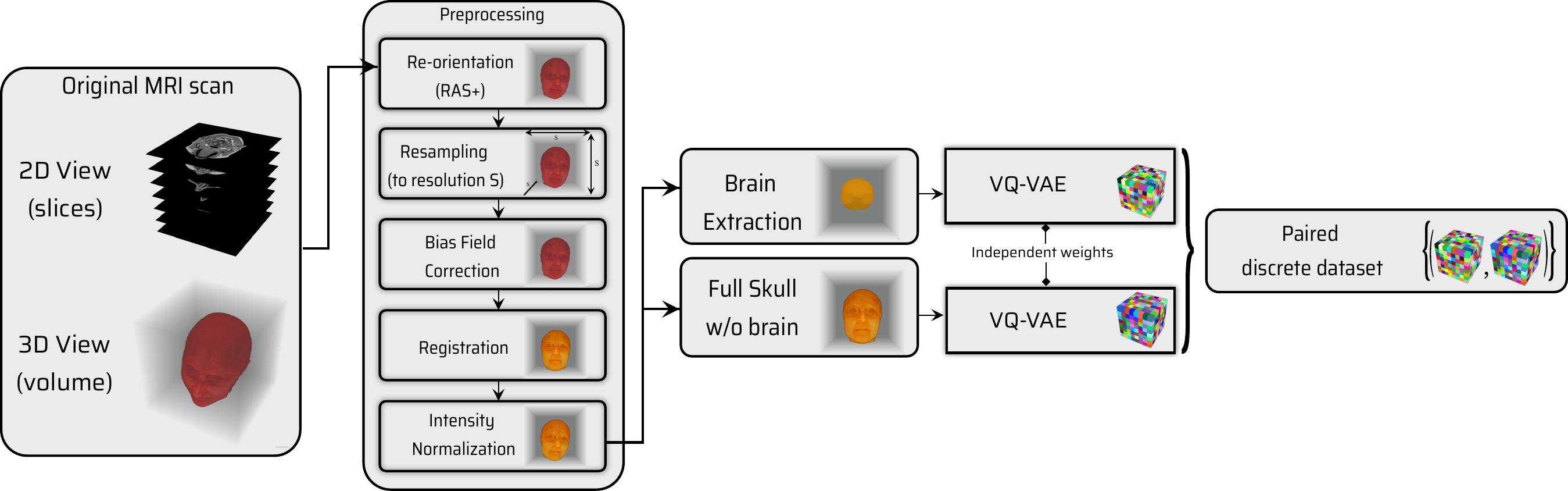}
    \caption{\textbf{Vector Quantization Stage.}  Following standard preprocessing we execute two independent stages: (i) We extract the brain of the scan, and inversely (ii) remove the brain from the full skull. Both representations are then \emph{compressed} independently by two \model{VQ-VAEs} into 3D integer volumes of much lower resolution. 
    }
    \label{fig:figPreprocessing}
\end{figure*}

\myparagraph{Image Synthesis.}

Generative models, including VAEs \cite{kingma2013auto}, GANs \cite{goodfellow2014generative}, diffusion models \cite{sohl2019denoising}, and masked autoencoders (MAEs) \cite{devlin2018bert,radford2018improving,radford2019language}, are extensively studied. MAEs excel by reconstructing images from degraded versions. Our work builds upon recent MAE advances like \model{MaskGIT} \cite{chang2022maskgit} and \model{Paella} \cite{rampas2022fast}.  To reduce computational requirements, we adopt MAEs over diffusion models, leveraging the former's faster inference.

%% file: 3_method.tex
\section{Method}

\myparagraph{Task Definition.} As in \cite{vandergoten2021conditional}, given a set of 3D scans $(X^{(i)})_{i=1,\ldots,N}$ following a data distribution $\mathcal{P}_X$ and having a resolution of $S^3$, we aim to find a mapping $Y = G_\Phi(\gamma(X))$ parameterized by $\Phi$ that de-identifies a \emph{raw} scan $X$ and yields $Y$. The purpose of the \emph{privacy transform} $\gamma(\cdot)$ is to provide $G_\Phi(\cdot)$ with a minimal blueprint to guide the skull synthesis without leaking information about the patient's facial features\footnote{As the skull is synthesized around the brain, $\gamma(\cdot)$ should at least contain a binary brain mask}. Fundamentally, $\gamma(X)$ should at least include a representation of the brain to inform the model about the metric constraints of the to-be-synthesized skull.

\begin{figure*}[t]
    \centering
    \includegraphics[width=\textwidth]{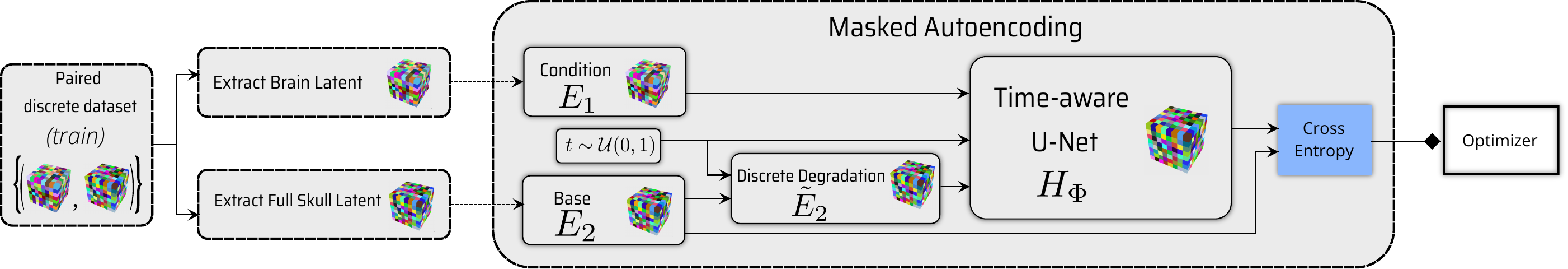}
    \caption{\textbf{Latent Modeling Stage (MAE).}  The two representations obtained in Fig \ref{fig:figPreprocessing} serve as \emph{conditioning} variable (brain) resp.\ to-be-degraded input (full skull) of the MAE. The MAE's task is to reverse the degradation.  }
    \label{fig:figLatent}
\end{figure*}

\begin{figure*}[t]
\centering
\includegraphics[width=\textwidth]{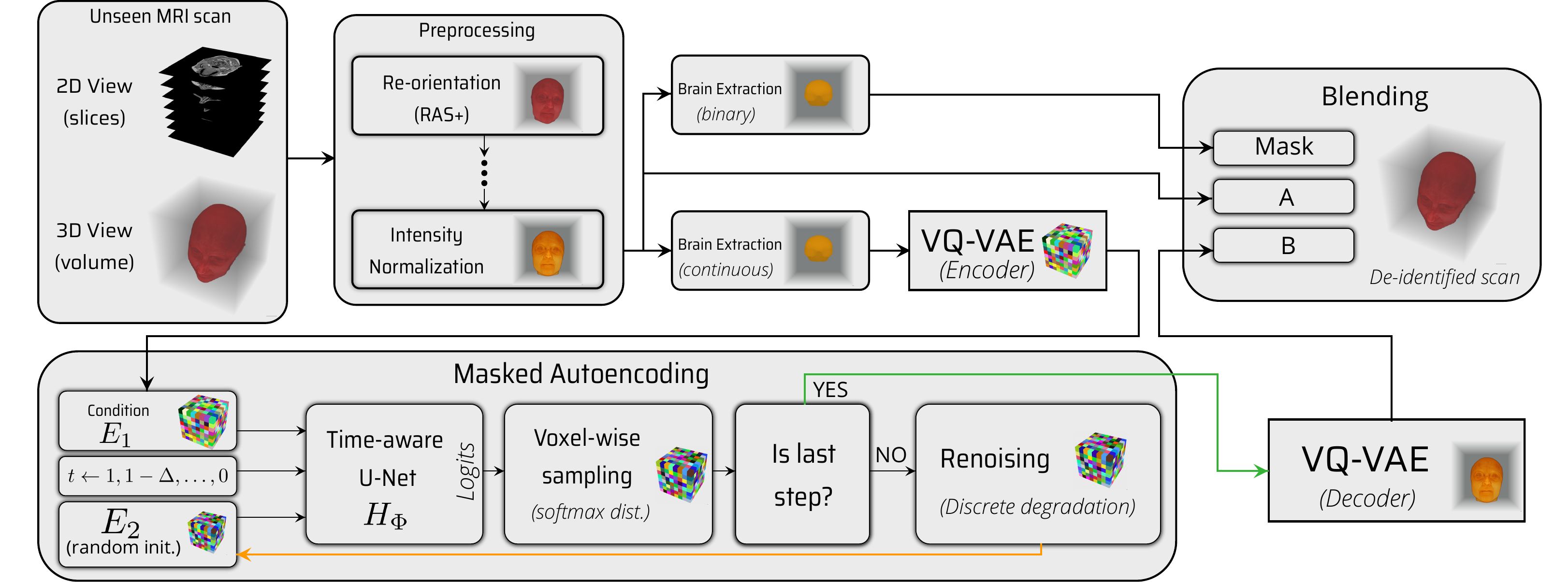}
\caption{\textbf{Test-time De-Identification.} We repeat the steps from Fig \ref{fig:figPreprocessing} to obtain the highly-compressed brain representation which is used as the \emph{condition} in the inference stage of the masked autoencoder. Starting from a randomly-initialized $\hat{X}$ the network refines its estimate of how a skull around the given brain could look like. The final \emph{de-identified} scan is obtained by \emph{blending} the original scan with  the last estimate $\hat{X}$ where the \emph{binary} brain representation acts as a \emph{mask}. This step ensures that the brain is preserved.}
\label{ref:figTestTime}
\end{figure*}

\myparagraph{Overview.}
Synthesizing (3D) MRI scans is a challenging endeavor in terms of memory as a volume contains a cubic number of voxels. Applying a \model{VQ-VAE} is therefore particularly promising as the number of voxels could in the most extreme case be reduced from $256^3$ to a mere $64^3$, reducing the overall number of voxels by a factor of 64. 
Our approach is depicted in Fig.~\ref{fig:figPreprocessing}).
One instance of the \model{VQ-VAE} is tasked to model the brains only, a second complementary instance models the full skulls \emph{without} the brain. 
Applying both trained instances on an MRI dataset produces a paired dataset where each item is a tuple of the two latent integer volumes coming from each \model{VQ-VAE} encoder. 
In a second step, we employ a \model{Paella}-style MAE that conditions on the previously derived latents from the brain and models the distribution of the latents pertaining to the full skulls (see Fig.~\ref{fig:figLatent}). 
Unseen MRI scans can then be de-identified by computing the latents associated to the brain and letting the MAE synthesize a realistic skull using the brain latent as conditioning variable (see Fig.~\ref{ref:figTestTime}).

\myparagraph{Vector Quantization Stage.}
In order to reduce the memory requirements of high-resolution MR imagery, we leverage the \model{VQ-VAE} framework to compress (parts of) MRI scans into latent codes.
As a preparatory step we use \model{ROBEX} \cite{iglesias2011robex} to compute the (binary) brain mask $B(X)$ of an MRI scan $X \in \mathcal{D}$. 
We need the brain representation for two reasons: it serves as a conditioning variable that informs the synthesis about the proportions of the skull, and we require it to later copy the original brain into the de-identified scan.

We then train a 3D \model{VQ-VAE} for each of the two representations $B(X) \odot X$ (\emph{brain}) and $\overline{B}(X) \odot X$ (\emph{skull}) \emph{in isolation} where $\odot$ denotes the Hadamard product and $\overline{B}(X)\ \hat{=}\ 1 - B(X)$. Both representations are complementary\footnote{By adding both one recovers $X$} to each other in that $B(X) \odot X$ contains solely the brain (\ie has non-zero brain intensities) without featuring any of the remaining parts (\ie zero non-brain intensities) whereas the inverted properties are true for $\overline{B}(X) \odot X$.

Training both models yields two encoder/decoder pairs $(e_1, d_1)$ and $(e_2, d_2)$. After training, we deploy the two encoders to translate the dataset of MRI scans $\mathcal{D}$ into a highly-compressed \emph{paired} dataset of integer codes $\mathcal{D}_e = \lbrace (e_1[B(X) \odot X)], e_2[\overline{B}(X)\odot X] \mid X \in \mathcal{D} \rbrace$. A depiction of this and the applied preprocessing can be found in Fig. \ref{fig:figPreprocessing}.

\myparagraph{Latent Modeling Stage.}
Having the compressed latent integer representations of $E_1 \hat{=} e_1[B(X) \odot X]$ (\emph{brain}) and $E_2 \hat{=} e_2[\overline{B}(X) \odot X]$ (\emph{skull}) now at our disposal, we describe a generative MAE $H_\Phi(E_2 | E_1)$ that predicts the latter (integer) skull representation while conditioning on the former (integer) brain representation.

Recall that both $E_1$ and $E_2$ are 3D volumes/grids of integers from the set  $\lbrace 0, \ldots, N_\text{CV} - 1\rbrace^{s \times s \times s}$. In order to model the distribution of $E_2$ the MAE methodology requires that one defines a (time-dependent) scheme to \emph{perturb} the integers in the $E_2$ representation. 
We decide to use \emph{Paella}-style \cite{rampas2022fast} perturbations, \ie a value $v \in \lbrace 0, \ldots, N_\text{CV} - 1 \rbrace $ is kept with probability $\alpha_t$ and resampled from the same set with a probability $1 - \alpha_t$, where the time $t \sim \mathcal{U}(0, 1)$ is sampled independently for each element in the batch.

\myparagraph{Sampling Stage.}
After having trained the MAE, we execute the following sampling steps: (i) Compute the brain mask $B(X')$, (ii) compute the integer representation $E_1 = e_1[B(X') \odot X']$, (iii) use $H_\Phi$ to sample $E_2$ using $E_1$ as a conditioning variable, and finally (iv), pass $E_2$ through the decoder $d_2(\cdot)$ to recover a skull that harmonizes well with the brain in $X'$.
To ensure that the de-identified scan accurately represents the real brain and not a hallucinated version thereof, we implement a simple blending scheme in order to copy the brain from the original $X$:
\begin{equation}
\label{eqn:EQ2}
    Y = (1 - B(X)) \odot d_2(\hat{E_2}) + B(X) \odot X
\end{equation}
where $\hat{E_2}$ denotes the latent as estimated by letting the MAE condition on $E_1 \hat{=} e_1[B(X) \odot X]$ (\ie the brain).

\begin{table*}[t]
    \centering
\resizebox{0.76\textwidth}{!}{
\begin{tabular}[t]{ccccc}
\toprule
\multicolumn{1}{c}{ } & \multicolumn{4}{c}{\emph{User-based}} \\
\cmidrule(l{3pt}r{3pt}){2-5} 
\multicolumn{1}{c}{} & \multicolumn{2}{c}{\dataset{ADNI}} & \multicolumn{2}{c}{\dataset{OASIS-3}} \\
\cmidrule(l{3pt}r{3pt}){2-3} \cmidrule(l{3pt}r{3pt}){4-5} 
 & $ 128^3 $ & $ 256^3 $ & $ 128^3 $ & $ 256^3 $ \\
\midrule
\model{BLACK}         & $ 20.17 \pm 25.45 $      & $ 20.30 \pm 26.33 $      & $ 18.37 \pm 10.90 $      & $ 18.54 \pm 11.08 $   \\
\model{BLURRED}       & $ 45.05 \pm 28.91 $      & $ 46.29 \pm 29.42 $      & $ 41.63 \pm 15.57 $      & $ 42.93 \pm 18.06 $   \\
\model{ORIGINAL}      & $ 55.33 \pm 30.70 $      & $ 58.93 \pm 29.24 $      & $ 61.86 \pm 15.00 $      & $ 59.27 \pm 16.94 $   \\
\model{MRI WATERSHED} & $ 19.03 \pm 25.31 $      & $ 21.48 \pm 27.25 $      & $ 22.56 \pm 13.11 $      & $ 22.20 \pm 14.41 $   \\
\hdashline
\model{DEFACE}        & $ 43.53 \pm 30.16 $      & $ 45.58 \pm 29.53 $      & $ 38.14 \pm 9.58 $       & $ 43.17 \pm 16.04 $   \\
\model{QUICKSHEAR}    & $ 38.51 \pm 30.18 $      & $ 39.81 \pm 30.43 $      & $ 40.70 \pm 16.68 $      & $ 35.85 \pm 13.78 $   \\
\model{FACE MASK v1}  & $ 48.55 \pm 30.49 $      & $ 50.24 \pm 30.54 $      & $ 50.23 \pm 17.39 $      & $ 52.68 \pm 14.32 $   \\
\model{FACE MASK v2}  & $ 38.05 \pm 28.79 $      & $ 43.60 \pm 29.90 $      & $ 33.02 \pm 17.93 $      & $ 35.85 \pm 15.00 $   \\
\model{CP-GAN}        & $ 28.46 \pm 27.95 $      & \xmark                   & $ 30.47 \pm 14.30 $      & \xmark                  \\
\model{CP-MAE}        & $ \bm{23.14 \pm 27.88} $ & $ \bm{25.91 \pm 27.65} $ & $ \bm{22.56 \pm 11.77} $ & $ \bm{23.41 \pm 15.27} $ \\
\bottomrule
\\
\end{tabular}
}
\resizebox{0.76\textwidth}{!}{
\begin{tabular}[t]{ccccc}
\toprule
\multicolumn{1}{c}{ } & \multicolumn{4}{c}{\emph{Model-based}} \\
\cmidrule(l{3pt}r{3pt}){2-5} 
\multicolumn{1}{c}{} & \multicolumn{2}{c}{\dataset{ADNI}} & \multicolumn{2}{c}{\dataset{OASIS-3}} \\
\cmidrule(l{3pt}r{3pt}){2-3} \cmidrule(l{3pt}r{3pt}){4-5}
 & $ 128^3 $ & $ 256^3 $ & $ 128^3 $ & $ 256^3 $ \\
\midrule
\model{BLACK}         &  $ 19.66 \pm 2.25 $      & $ 18.75 \pm 2.08 $      & $ 19.86 \pm 0.89 $      & $ 19.98 \pm 1.19 $\\
\model{BLURRED}       &  $ 87.46 \pm 7.65 $      & $ 86.54 \pm 8.61 $      & $ 97.05 \pm 2.33 $      & $ 97.31 \pm 1.67 $\\
\model{ORIGINAL}      &  $ 100.00 \pm 0.00 $     & $ 100.00 \pm 0.00 $     & $ 100.00 \pm 0.00 $     & $ 100.00 \pm 0.00 $\\
\model{MRI WATERSHED} &  $ 44.75 \pm 4.09 $      & $ 47.03 \pm 5.06 $      & $ 67.76 \pm 4.39 $      & $ 67.50 \pm 7.39 $\\
\hdashline
\model{DEFACE}        &  $ 99.27 \pm 0.86 $      & $ 99.22 \pm 1.51 $      & $ 99.78 \pm 0.36 $      & $ 99.85 \pm 0.20 $\\
\model{QUICKSHEAR}    &  $ 98.79 \pm 0.87 $      & $ 95.66 \pm 1.98 $      & $ 99.81 \pm 0.21 $      & $ 99.83 \pm 0.20 $\\
\model{FACE MASK v1}  &  $ 96.31 \pm 2.71 $      & $ 98.75 \pm 1.59 $      & $ 99.65 \pm 0.50 $      & $ 99.71 \pm 0.23 $\\
\model{FACE MASK v2}  &  $ 94.42 \pm 5.11 $      & $ 98.06 \pm 1.20 $      & $ 99.78 \pm 0.31 $      & $ 99.63 \pm 0.29 $\\
\model{CP-GAN}        &  $ 56.11 \pm 5.05 $      & \xmark                  & $ 56.40 \pm 2.89 $      & \xmark \\
\model{CP-MAE}        &  $ \bm{39.91 \pm 9.49} $ & $ \bm{41.74 \pm 6.91} $ & $ \bm{48.82 \pm 4.32} $ & $ \bm{58.19 \pm 4.04} $\\
\bottomrule
\\
\end{tabular}

}
    \caption{\textbf{De-Identification Quality.} We compare \model{CP-MAE} against traditional methods, \model{CP-GAN} and four control methods (above "- -" line) in terms of their de-identification capabilities \wrt a user-based (top) and a model-based (bottom) task. Both scenarios involve finding the correct option among five alternatives (of which four belong to a different subject) and the showcased values correspond to the percentage of correct guesses ($\pm$ s.d.). ("\xmark" indicates an unsupported resolution) }
    \label{tab:de-identification}
\end{table*}

%% file: 4_experiments.tex
\section{Experiments}

We evaluate our MRI de-identification method's effectiveness and diagnostic value retention through privacy protection and downstream task performance assessments. 
This includes the segmentation tasks FIRST \cite{fslFirst} and FASTSURFER \cite{Henschel2019FastSurferA}) to gauge the impact of de-identification on real-world applications, along with user and model-based studies to determine re-identification difficulty.

\myparagraph{Datasets \& Implementation}
Our study utilizes the \dataset{ADNI} \cite{Weiner2017,Wyman2013} (2,172 scans) and \dataset{OASIS-3} \cite{LaMontagne2019.12.13.19014902} (2,556 scans) datasets, with a per-patient 80-20 train-test split. For training our network, we leverage two NVIDIA RTX 4090 GPUs, each equipped with 24 GiB of GPU memory. Our implementation relies on the PyTorch framework \cite{pytorch}. To optimize memory consumption, we utilize the \emph{inductor} backend for model compilation, which has proven to be highly advantageous. 
\myparagraph{Benchmarks.}
Our study compares six de-identification methods: three removal-based (\model{QUICKSHEAR}, \model{FACE MASK v1}, \model{DEFACE}), \model{MRI WATERSHED} (removes non-brain tissue), \model{FACE MASK v2} (an update), and \model{CP-GAN} (a remodeling-based method). Applied to $128^3$ and $256^3$ resolution images (except \model{CP-GAN}, limited to $128^3$).

\myparagraph{De-Identification Quality: User-based Study.}
In our Amazon Mechanical Turk study appearing in Table \ref{tab:de-identification}, we assessed the de-identification methods' resilience by having participants match original 3D scans with de-identified versions from five choices, using side-profile views to enhance feature discernibility. 
Across 22,000 responses, CP-MAE excelled in de-identification, outdoing traditional methods by 15-25\

\begin{figure*}[t]
    \centering
        \resizebox{\linewidth}{!}{
    \input{figs/experiments/subcortical/plot_first_fastsurfer.tikz}
    }
    \caption{\textbf{Downstream Tasks: Subcortical segmentation.} We analyze to which extent de-identification affects the quality of subcortical segmentation methods. The depicted values are the \emph{class-averaged} Dice scores over $15$ classes for  FIRST and resp.\ $78$ classes for FASTSURFER. To increase visual discernability we excluded \model{MRI WATERSHED} (avg.\ $\approx 0.21$). Higher values are preferable. }
    \label{fig:first_fastsurfer_results}
\end{figure*}
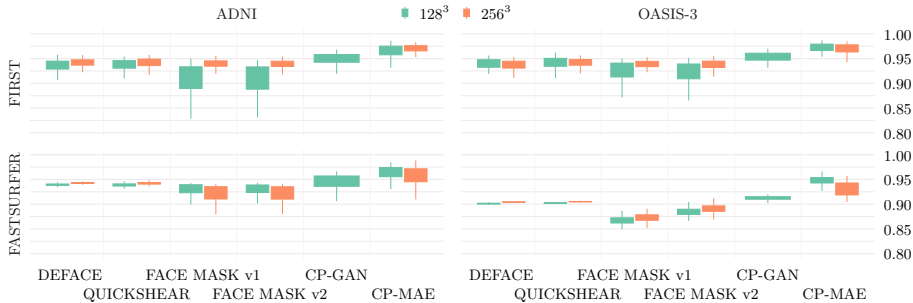

\myparagraph{De-Identification Quality: Model-based Study.}
We tested the de-identification models' robustness using a neural network to compare original and de-identified images, similar to the CP-GAN study but with a more powerful similarity-quantification network. 
Said Siamese network was trained to recognize patient-matched images, utilizing ResNet-18 for embeddings and computing distances with the Triplet Margin loss. 
Then, for each test fold, we present a patient's original scan and its corresponding de-identified version (using method $m$) along with four incorrect options randomly chosen $m$-renderings from different patients. 
The network attempts to re-identify by selecting the option, $y$, minimizing the Euclidean distance to $\tilde{h}(x)$.
The results in Table \ref{tab:de-identification}, show CP-MAE with the highest resistance to re-identification across the board.

\myparagraph{Effect of De-Identification on Medical Analyses.} 
To ensure de-identification does not hinder medical analysis, we tested its impact using the subcortical brain segmentation algorithms FIRST \cite{fslFirst}, and FASTSURFER \cite{Henschel2019FastSurferA}, for an even more fine-grained segmentation. Original and de-identified scan segmentations are compared with the Dice coefficient. CP-MAE maintained analysis integrity, outperforming others with minimal effect on FIRST and FASTSURFER outcomes. In contrast, \model{MRI WATERSHED} significantly disrupted downstream analyses, underscoring the shortcomings of methods that simply retain the brain and discard everything else.

%% file: figs/experiments/subcortical/plot_first_fastsurfer.tikz
\begin{tikzpicture}[x=1pt,y=1pt]
\definecolor{fillColor}{RGB}{255,255,255}
\path[use as bounding box,fill=fillColor,fill opacity=0.00] (0,0) rectangle (4336.20,1445.40);
\begin{scope}
\path[clip] (150.65,778.98) rectangle (2107.15,1294.75);
\definecolor{drawColor}{gray}{0.92}

\path[draw=drawColor,line width= 2.6pt,line join=round] (150.65,861.03) --
	(2107.15,861.03);

\path[draw=drawColor,line width= 2.6pt,line join=round] (150.65,978.25) --
	(2107.15,978.25);

\path[draw=drawColor,line width= 2.6pt,line join=round] (150.65,1095.47) --
	(2107.15,1095.47);

\path[draw=drawColor,line width= 2.6pt,line join=round] (150.65,1212.70) --
	(2107.15,1212.70);

\path[draw=drawColor,line width= 5.2pt,line join=round] (150.65,802.42) --
	(2107.15,802.42);

\path[draw=drawColor,line width= 5.2pt,line join=round] (150.65,919.64) --
	(2107.15,919.64);

\path[draw=drawColor,line width= 5.2pt,line join=round] (150.65,1036.86) --
	(2107.15,1036.86);

\path[draw=drawColor,line width= 5.2pt,line join=round] (150.65,1154.09) --
	(2107.15,1154.09);

\path[draw=drawColor,line width= 5.2pt,line join=round] (150.65,1271.31) --
	(2107.15,1271.31);
\definecolor{drawColor}{RGB}{102,194,165}

\path[draw=drawColor,line width= 0.6pt,line join=round] (280.82,1142.16) -- (280.82,1172.28);

\path[draw=drawColor,line width= 0.6pt,line join=round] (280.82,1104.59) -- (280.82,1051.53);
\definecolor{fillColor}{RGB}{102,194,165}

\path[draw=drawColor,line width= 0.6pt,fill=fillColor] (227.57,1142.16) --
	(227.57,1104.59) --
	(334.07,1104.59) --
	(334.07,1142.16) --
	(227.57,1142.16) --
	cycle;

\path[draw=drawColor,line width= 0.6pt,line join=round] (596.38,1144.56) -- (596.38,1163.50);

\path[draw=drawColor,line width= 0.6pt,line join=round] (596.38,1108.76) -- (596.38,1059.78);

\path[draw=drawColor,line width= 0.6pt,fill=fillColor] (543.13,1144.56) --
	(543.13,1108.76) --
	(649.64,1108.76) --
	(649.64,1144.56) --
	(543.13,1144.56) --
	cycle;

\path[draw=drawColor,line width= 0.6pt,line join=round] (911.95,1115.26) -- (911.95,1154.17);

\path[draw=drawColor,line width= 0.6pt,line join=round] (911.95,1012.58) -- (911.95,869.52);

\path[draw=drawColor,line width= 0.6pt,fill=fillColor] (858.70,1115.26) --
	(858.70,1012.58) --
	(965.20,1012.58) --
	(965.20,1115.26) --
	(858.70,1115.26) --
	cycle;

\path[draw=drawColor,line width= 0.6pt,line join=round] (1227.51,1114.65) -- (1227.51,1147.77);

\path[draw=drawColor,line width= 0.6pt,line join=round] (1227.51,1009.21) -- (1227.51,874.50);

\path[draw=drawColor,line width= 0.6pt,fill=fillColor] (1174.26,1114.65) --
	(1174.26,1009.21) --
	(1280.77,1009.21) --
	(1280.77,1114.65) --
	(1174.26,1114.65) --
	cycle;

\path[draw=drawColor,line width= 0.6pt,line join=round] (1602.25,1173.20) -- (1602.25,1196.67);

\path[draw=drawColor,line width= 0.6pt,line join=round] (1602.25,1136.75) -- (1602.25,1082.45);

\path[draw=drawColor,line width= 0.6pt,fill=fillColor] (1495.74,1173.20) --
	(1495.74,1136.75) --
	(1708.75,1136.75) --
	(1708.75,1173.20) --
	(1495.74,1173.20) --
	cycle;

\path[draw=drawColor,line width= 0.6pt,line join=round] (1858.64,1213.21) -- (1858.64,1238.69);

\path[draw=drawColor,line width= 0.6pt,line join=round] (1858.64,1172.12) -- (1858.64,1111.13);

\path[draw=drawColor,line width= 0.6pt,fill=fillColor] (1805.39,1213.21) --
	(1805.39,1172.12) --
	(1911.90,1172.12) --
	(1911.90,1213.21) --
	(1805.39,1213.21) --
	cycle;
\definecolor{drawColor}{RGB}{252,141,98}

\path[draw=drawColor,line width= 0.6pt,line join=round] (399.16,1148.37) -- (399.16,1170.29);

\path[draw=drawColor,line width= 0.6pt,line join=round] (399.16,1123.07) -- (399.16,1089.08);
\definecolor{fillColor}{RGB}{252,141,98}

\path[draw=drawColor,line width= 0.6pt,fill=fillColor] (345.91,1148.37) --
	(345.91,1123.07) --
	(452.41,1123.07) --
	(452.41,1148.37) --
	(345.91,1148.37) --
	cycle;

\path[draw=drawColor,line width= 0.6pt,line join=round] (714.72,1151.89) -- (714.72,1171.96);

\path[draw=drawColor,line width= 0.6pt,line join=round] (714.72,1121.02) -- (714.72,1077.07);

\path[draw=drawColor,line width= 0.6pt,fill=fillColor] (661.47,1151.89) --
	(661.47,1121.02) --
	(767.97,1121.02) --
	(767.97,1151.89) --
	(661.47,1151.89) --
	cycle;

\path[draw=drawColor,line width= 0.6pt,line join=round] (1030.29,1143.29) -- (1030.29,1166.67);

\path[draw=drawColor,line width= 0.6pt,line join=round] (1030.29,1118.08) -- (1030.29,1081.73);

\path[draw=drawColor,line width= 0.6pt,fill=fillColor] (977.04,1143.29) --
	(977.04,1118.08) --
	(1083.54,1118.08) --
	(1083.54,1143.29) --
	(977.04,1143.29) --
	cycle;

\path[draw=drawColor,line width= 0.6pt,line join=round] (1345.85,1141.82) -- (1345.85,1162.87);

\path[draw=drawColor,line width= 0.6pt,line join=round] (1345.85,1116.76) -- (1345.85,1079.72);

\path[draw=drawColor,line width= 0.6pt,fill=fillColor] (1292.60,1141.82) --
	(1292.60,1116.76) --
	(1399.10,1116.76) --
	(1399.10,1141.82) --
	(1292.60,1141.82) --
	cycle;

\path[draw=drawColor,line width= 0.6pt,line join=round] (1976.98,1215.45) -- (1976.98,1232.23);

\path[draw=drawColor,line width= 0.6pt,line join=round] (1976.98,1190.66) -- (1976.98,1161.16);

\path[draw=drawColor,line width= 0.6pt,fill=fillColor] (1923.73,1215.45) --
	(1923.73,1190.66) --
	(2030.23,1190.66) --
	(2030.23,1215.45) --
	(1923.73,1215.45) --
	cycle;
\definecolor{drawColor}{RGB}{255,240,245}

\path[draw=drawColor,line width= 0.0pt,line join=round] (497.77,778.98) -- (497.77,1294.75);

\path[draw=drawColor,line width= 0.0pt,line join=round] (813.34,778.98) -- (813.34,1294.75);

\path[draw=drawColor,line width= 0.0pt,line join=round] (1128.90,778.98) -- (1128.90,1294.75);

\path[draw=drawColor,line width= 0.0pt,line join=round] (1444.47,778.98) -- (1444.47,1294.75);

\path[draw=drawColor,line width= 0.0pt,line join=round] (1760.03,778.98) -- (1760.03,1294.75);
\end{scope}
\begin{scope}
\path[clip] (150.65,206.30) rectangle (2107.15,722.07);
\definecolor{drawColor}{gray}{0.92}

\path[draw=drawColor,line width= 2.6pt,line join=round] (150.65,288.35) --
	(2107.15,288.35);

\path[draw=drawColor,line width= 2.6pt,line join=round] (150.65,405.57) --
	(2107.15,405.57);

\path[draw=drawColor,line width= 2.6pt,line join=round] (150.65,522.80) --
	(2107.15,522.80);

\path[draw=drawColor,line width= 2.6pt,line join=round] (150.65,640.02) --
	(2107.15,640.02);

\path[draw=drawColor,line width= 5.2pt,line join=round] (150.65,229.74) --
	(2107.15,229.74);

\path[draw=drawColor,line width= 5.2pt,line join=round] (150.65,346.96) --
	(2107.15,346.96);

\path[draw=drawColor,line width= 5.2pt,line join=round] (150.65,464.19) --
	(2107.15,464.19);

\path[draw=drawColor,line width= 5.2pt,line join=round] (150.65,581.41) --
	(2107.15,581.41);

\path[draw=drawColor,line width= 5.2pt,line join=round] (150.65,698.63) --
	(2107.15,698.63);
\definecolor{drawColor}{RGB}{102,194,165}

\path[draw=drawColor,line width= 0.6pt,line join=round] (280.82,559.91) -- (280.82,568.31);

\path[draw=drawColor,line width= 0.6pt,line join=round] (280.82,553.36) -- (280.82,543.93);
\definecolor{fillColor}{RGB}{102,194,165}

\path[draw=drawColor,line width= 0.6pt,fill=fillColor] (227.57,559.91) --
	(227.57,553.36) --
	(334.07,553.36) --
	(334.07,559.91) --
	(227.57,559.91) --
	cycle;

\path[draw=drawColor,line width= 0.6pt,line join=round] (596.38,560.37) -- (596.38,573.92);

\path[draw=drawColor,line width= 0.6pt,line join=round] (596.38,550.19) -- (596.38,536.87);

\path[draw=drawColor,line width= 0.6pt,fill=fillColor] (543.13,560.37) --
	(543.13,550.19) --
	(649.64,550.19) --
	(649.64,560.37) --
	(543.13,560.37) --
	cycle;

\path[draw=drawColor,line width= 0.6pt,line join=round] (911.95,556.48) -- (911.95,563.67);

\path[draw=drawColor,line width= 0.6pt,line join=round] (911.95,518.32) -- (911.95,462.86);

\path[draw=drawColor,line width= 0.6pt,fill=fillColor] (858.70,556.48) --
	(858.70,518.32) --
	(965.20,518.32) --
	(965.20,556.48) --
	(858.70,556.48) --
	cycle;

\path[draw=drawColor,line width= 0.6pt,line join=round] (1227.51,555.23) -- (1227.51,564.85);

\path[draw=drawColor,line width= 0.6pt,line join=round] (1227.51,519.49) -- (1227.51,468.47);

\path[draw=drawColor,line width= 0.6pt,fill=fillColor] (1174.26,555.23) --
	(1174.26,519.49) --
	(1280.77,519.49) --
	(1280.77,555.23) --
	(1174.26,555.23) --
	cycle;

\path[draw=drawColor,line width= 0.6pt,line join=round] (1602.25,597.82) -- (1602.25,619.33);

\path[draw=drawColor,line width= 0.6pt,line join=round] (1602.25,548.93) -- (1602.25,478.82);

\path[draw=drawColor,line width= 0.6pt,fill=fillColor] (1495.74,597.82) --
	(1495.74,548.93) --
	(1708.75,548.93) --
	(1708.75,597.82) --
	(1495.74,597.82) --
	cycle;

\path[draw=drawColor,line width= 0.6pt,line join=round] (1858.64,637.68) -- (1858.64,662.55);

\path[draw=drawColor,line width= 0.6pt,line join=round] (1858.64,594.26) -- (1858.64,535.49);

\path[draw=drawColor,line width= 0.6pt,fill=fillColor] (1805.39,637.68) --
	(1805.39,594.26) --
	(1911.90,594.26) --
	(1911.90,637.68) --
	(1805.39,637.68) --
	cycle;
\definecolor{drawColor}{RGB}{252,141,98}

\path[draw=drawColor,line width= 0.6pt,line join=round] (399.16,566.65) -- (399.16,572.59);

\path[draw=drawColor,line width= 0.6pt,line join=round] (399.16,562.42) -- (399.16,556.65);
\definecolor{fillColor}{RGB}{252,141,98}

\path[draw=drawColor,line width= 0.6pt,fill=fillColor] (345.91,566.65) --
	(345.91,562.42) --
	(452.41,562.42) --
	(452.41,566.65) --
	(345.91,566.65) --
	cycle;

\path[draw=drawColor,line width= 0.6pt,line join=round] (714.72,566.96) -- (714.72,577.14);

\path[draw=drawColor,line width= 0.6pt,line join=round] (714.72,559.36) -- (714.72,548.43);

\path[draw=drawColor,line width= 0.6pt,fill=fillColor] (661.47,566.96) --
	(661.47,559.36) --
	(767.97,559.36) --
	(767.97,566.96) --
	(661.47,566.96) --
	cycle;

\path[draw=drawColor,line width= 0.6pt,line join=round] (1030.29,547.20) -- (1030.29,561.00);

\path[draw=drawColor,line width= 0.6pt,line join=round] (1030.29,488.77) -- (1030.29,415.32);

\path[draw=drawColor,line width= 0.6pt,fill=fillColor] (977.04,547.20) --
	(977.04,488.77) --
	(1083.54,488.77) --
	(1083.54,547.20) --
	(977.04,547.20) --
	cycle;

\path[draw=drawColor,line width= 0.6pt,line join=round] (1345.85,546.76) -- (1345.85,560.84);

\path[draw=drawColor,line width= 0.6pt,line join=round] (1345.85,488.25) -- (1345.85,418.10);

\path[draw=drawColor,line width= 0.6pt,fill=fillColor] (1292.60,546.76) --
	(1292.60,488.25) --
	(1399.10,488.25) --
	(1399.10,546.76) --
	(1292.60,546.76) --
	cycle;

\path[draw=drawColor,line width= 0.6pt,line join=round] (1976.98,631.84) -- (1976.98,672.16);

\path[draw=drawColor,line width= 0.6pt,line join=round] (1976.98,570.24) -- (1976.98,486.71);

\path[draw=drawColor,line width= 0.6pt,fill=fillColor] (1923.73,631.84) --
	(1923.73,570.24) --
	(2030.23,570.24) --
	(2030.23,631.84) --
	(1923.73,631.84) --
	cycle;
\definecolor{drawColor}{RGB}{255,240,245}

\path[draw=drawColor,line width= 0.0pt,line join=round] (497.77,206.30) -- (497.77,722.07);

\path[draw=drawColor,line width= 0.0pt,line join=round] (813.34,206.30) -- (813.34,722.07);

\path[draw=drawColor,line width= 0.0pt,line join=round] (1128.90,206.30) -- (1128.90,722.07);

\path[draw=drawColor,line width= 0.0pt,line join=round] (1444.47,206.30) -- (1444.47,722.07);

\path[draw=drawColor,line width= 0.0pt,line join=round] (1760.03,206.30) -- (1760.03,722.07);
\end{scope}
\begin{scope}
\path[clip] (2192.51,778.98) rectangle (4149.01,1294.75);
\definecolor{drawColor}{gray}{0.92}

\path[draw=drawColor,line width= 2.6pt,line join=round] (2192.51,861.03) --
	(4149.01,861.03);

\path[draw=drawColor,line width= 2.6pt,line join=round] (2192.51,978.25) --
	(4149.01,978.25);

\path[draw=drawColor,line width= 2.6pt,line join=round] (2192.51,1095.47) --
	(4149.01,1095.47);

\path[draw=drawColor,line width= 2.6pt,line join=round] (2192.51,1212.70) --
	(4149.01,1212.70);

\path[draw=drawColor,line width= 5.2pt,line join=round] (2192.51,802.42) --
	(4149.01,802.42);

\path[draw=drawColor,line width= 5.2pt,line join=round] (2192.51,919.64) --
	(4149.01,919.64);

\path[draw=drawColor,line width= 5.2pt,line join=round] (2192.51,1036.86) --
	(4149.01,1036.86);

\path[draw=drawColor,line width= 5.2pt,line join=round] (2192.51,1154.09) --
	(4149.01,1154.09);

\path[draw=drawColor,line width= 5.2pt,line join=round] (2192.51,1271.31) --
	(4149.01,1271.31);
\definecolor{drawColor}{RGB}{102,194,165}

\path[draw=drawColor,line width= 0.6pt,line join=round] (2322.68,1148.93) -- (2322.68,1168.76);

\path[draw=drawColor,line width= 0.6pt,line join=round] (2322.68,1113.05) -- (2322.68,1081.12);
\definecolor{fillColor}{RGB}{102,194,165}

\path[draw=drawColor,line width= 0.6pt,fill=fillColor] (2269.43,1148.93) --
	(2269.43,1113.05) --
	(2375.93,1113.05) --
	(2375.93,1148.93) --
	(2269.43,1148.93) --
	cycle;

\path[draw=drawColor,line width= 0.6pt,line join=round] (2638.25,1155.39) -- (2638.25,1182.90);

\path[draw=drawColor,line width= 0.6pt,line join=round] (2638.25,1117.70) -- (2638.25,1062.57);

\path[draw=drawColor,line width= 0.6pt,fill=fillColor] (2584.99,1155.39) --
	(2584.99,1117.70) --
	(2691.50,1117.70) --
	(2691.50,1155.39) --
	(2584.99,1155.39) --
	cycle;

\path[draw=drawColor,line width= 0.6pt,line join=round] (2953.81,1132.80) -- (2953.81,1154.72);

\path[draw=drawColor,line width= 0.6pt,line join=round] (2953.81,1067.02) -- (2953.81,969.31);

\path[draw=drawColor,line width= 0.6pt,fill=fillColor] (2900.56,1132.80) --
	(2900.56,1067.02) --
	(3007.06,1067.02) --
	(3007.06,1132.80) --
	(2900.56,1132.80) --
	cycle;

\path[draw=drawColor,line width= 0.6pt,line join=round] (3269.38,1128.58) -- (3269.38,1157.48);

\path[draw=drawColor,line width= 0.6pt,line join=round] (3269.38,1058.45) -- (3269.38,955.72);

\path[draw=drawColor,line width= 0.6pt,fill=fillColor] (3216.12,1128.58) --
	(3216.12,1058.45) --
	(3322.63,1058.45) --
	(3322.63,1128.58) --
	(3216.12,1128.58) --
	cycle;

\path[draw=drawColor,line width= 0.6pt,line join=round] (3644.11,1179.76) -- (3644.11,1201.67);

\path[draw=drawColor,line width= 0.6pt,line join=round] (3644.11,1146.91) -- (3644.11,1111.30);

\path[draw=drawColor,line width= 0.6pt,fill=fillColor] (3537.61,1179.76) --
	(3537.61,1146.91) --
	(3750.61,1146.91) --
	(3750.61,1179.76) --
	(3537.61,1179.76) --
	cycle;

\path[draw=drawColor,line width= 0.6pt,line join=round] (3900.51,1222.87) -- (3900.51,1241.17);

\path[draw=drawColor,line width= 0.6pt,line join=round] (3900.51,1192.30) -- (3900.51,1162.94);

\path[draw=drawColor,line width= 0.6pt,fill=fillColor] (3847.25,1222.87) --
	(3847.25,1192.30) --
	(3953.76,1192.30) --
	(3953.76,1222.87) --
	(3847.25,1222.87) --
	cycle;
\definecolor{drawColor}{RGB}{252,141,98}

\path[draw=drawColor,line width= 0.6pt,line join=round] (2441.02,1141.88) -- (2441.02,1161.32);

\path[draw=drawColor,line width= 0.6pt,line join=round] (2441.02,1109.16) -- (2441.02,1062.84);
\definecolor{fillColor}{RGB}{252,141,98}

\path[draw=drawColor,line width= 0.6pt,fill=fillColor] (2387.77,1141.88) --
	(2387.77,1109.16) --
	(2494.27,1109.16) --
	(2494.27,1141.88) --
	(2387.77,1141.88) --
	cycle;

\path[draw=drawColor,line width= 0.6pt,line join=round] (2756.58,1149.54) -- (2756.58,1169.23);

\path[draw=drawColor,line width= 0.6pt,line join=round] (2756.58,1122.66) -- (2756.58,1083.76);

\path[draw=drawColor,line width= 0.6pt,fill=fillColor] (2703.33,1149.54) --
	(2703.33,1122.66) --
	(2809.83,1122.66) --
	(2809.83,1149.54) --
	(2703.33,1149.54) --
	cycle;

\path[draw=drawColor,line width= 0.6pt,line join=round] (3072.15,1141.00) -- (3072.15,1161.04);

\path[draw=drawColor,line width= 0.6pt,line join=round] (3072.15,1115.72) -- (3072.15,1089.42);

\path[draw=drawColor,line width= 0.6pt,fill=fillColor] (3018.90,1141.00) --
	(3018.90,1115.72) --
	(3125.40,1115.72) --
	(3125.40,1141.00) --
	(3018.90,1141.00) --
	cycle;

\path[draw=drawColor,line width= 0.6pt,line join=round] (3387.71,1142.24) -- (3387.71,1165.30);

\path[draw=drawColor,line width= 0.6pt,line join=round] (3387.71,1111.67) -- (3387.71,1067.92);

\path[draw=drawColor,line width= 0.6pt,fill=fillColor] (3334.46,1142.24) --
	(3334.46,1111.67) --
	(3440.96,1111.67) --
	(3440.96,1142.24) --
	(3334.46,1142.24) --
	cycle;

\path[draw=drawColor,line width= 0.6pt,line join=round] (4018.84,1219.36) -- (4018.84,1235.88);

\path[draw=drawColor,line width= 0.6pt,line join=round] (4018.84,1185.60) -- (4018.84,1136.05);

\path[draw=drawColor,line width= 0.6pt,fill=fillColor] (3965.59,1219.36) --
	(3965.59,1185.60) --
	(4072.09,1185.60) --
	(4072.09,1219.36) --
	(3965.59,1219.36) --
	cycle;
\definecolor{drawColor}{RGB}{255,240,245}

\path[draw=drawColor,line width= 0.0pt,line join=round] (2539.63,778.98) -- (2539.63,1294.75);

\path[draw=drawColor,line width= 0.0pt,line join=round] (2855.20,778.98) -- (2855.20,1294.75);

\path[draw=drawColor,line width= 0.0pt,line join=round] (3170.76,778.98) -- (3170.76,1294.75);

\path[draw=drawColor,line width= 0.0pt,line join=round] (3486.33,778.98) -- (3486.33,1294.75);

\path[draw=drawColor,line width= 0.0pt,line join=round] (3801.89,778.98) -- (3801.89,1294.75);
\end{scope}
\begin{scope}
\path[clip] (2192.51,206.30) rectangle (4149.01,722.07);
\definecolor{drawColor}{gray}{0.92}

\path[draw=drawColor,line width= 2.6pt,line join=round] (2192.51,288.35) --
	(4149.01,288.35);

\path[draw=drawColor,line width= 2.6pt,line join=round] (2192.51,405.57) --
	(4149.01,405.57);

\path[draw=drawColor,line width= 2.6pt,line join=round] (2192.51,522.80) --
	(4149.01,522.80);

\path[draw=drawColor,line width= 2.6pt,line join=round] (2192.51,640.02) --
	(4149.01,640.02);

\path[draw=drawColor,line width= 5.2pt,line join=round] (2192.51,229.74) --
	(4149.01,229.74);

\path[draw=drawColor,line width= 5.2pt,line join=round] (2192.51,346.96) --
	(4149.01,346.96);

\path[draw=drawColor,line width= 5.2pt,line join=round] (2192.51,464.19) --
	(4149.01,464.19);

\path[draw=drawColor,line width= 5.2pt,line join=round] (2192.51,581.41) --
	(4149.01,581.41);

\path[draw=drawColor,line width= 5.2pt,line join=round] (2192.51,698.63) --
	(4149.01,698.63);
\definecolor{drawColor}{RGB}{102,194,165}

\path[draw=drawColor,line width= 0.6pt,line join=round] (2322.68,468.76) -- (2322.68,473.21);

\path[draw=drawColor,line width= 0.6pt,line join=round] (2322.68,464.52) -- (2322.68,460.65);
\definecolor{fillColor}{RGB}{102,194,165}

\path[draw=drawColor,line width= 0.6pt,fill=fillColor] (2269.43,468.76) --
	(2269.43,464.52) --
	(2375.93,464.52) --
	(2375.93,468.76) --
	(2269.43,468.76) --
	cycle;

\path[draw=drawColor,line width= 0.6pt,line join=round] (2638.25,470.94) -- (2638.25,474.77);

\path[draw=drawColor,line width= 0.6pt,line join=round] (2638.25,468.35) -- (2638.25,464.53);

\path[draw=drawColor,line width= 0.6pt,fill=fillColor] (2584.99,470.94) --
	(2584.99,468.35) --
	(2691.50,468.35) --
	(2691.50,470.94) --
	(2584.99,470.94) --
	cycle;

\path[draw=drawColor,line width= 0.6pt,line join=round] (2953.81,399.53) -- (2953.81,432.91);

\path[draw=drawColor,line width= 0.6pt,line join=round] (2953.81,374.84) -- (2953.81,344.15);

\path[draw=drawColor,line width= 0.6pt,fill=fillColor] (2900.56,399.53) --
	(2900.56,374.84) --
	(3007.06,374.84) --
	(3007.06,399.53) --
	(2900.56,399.53) --
	cycle;

\path[draw=drawColor,line width= 0.6pt,line join=round] (3269.38,440.43) -- (3269.38,474.95);

\path[draw=drawColor,line width= 0.6pt,line join=round] (3269.38,415.71) -- (3269.38,384.60);

\path[draw=drawColor,line width= 0.6pt,fill=fillColor] (3216.12,440.43) --
	(3216.12,415.71) --
	(3322.63,415.71) --
	(3322.63,440.43) --
	(3216.12,440.43) --
	cycle;

\path[draw=drawColor,line width= 0.6pt,line join=round] (3644.11,499.85) -- (3644.11,511.13);

\path[draw=drawColor,line width= 0.6pt,line join=round] (3644.11,487.68) -- (3644.11,469.59);

\path[draw=drawColor,line width= 0.6pt,fill=fillColor] (3537.61,499.85) --
	(3537.61,487.68) --
	(3750.61,487.68) --
	(3750.61,499.85) --
	(3537.61,499.85) --
	cycle;

\path[draw=drawColor,line width= 0.6pt,line join=round] (3900.51,590.31) -- (3900.51,618.66);

\path[draw=drawColor,line width= 0.6pt,line join=round] (3900.51,564.64) -- (3900.51,526.55);

\path[draw=drawColor,line width= 0.6pt,fill=fillColor] (3847.25,590.31) --
	(3847.25,564.64) --
	(3953.76,564.64) --
	(3953.76,590.31) --
	(3847.25,590.31) --
	cycle;
\definecolor{drawColor}{RGB}{252,141,98}

\path[draw=drawColor,line width= 0.6pt,line join=round] (2441.02,475.38) -- (2441.02,478.55);

\path[draw=drawColor,line width= 0.6pt,line join=round] (2441.02,472.57) -- (2441.02,469.19);
\definecolor{fillColor}{RGB}{252,141,98}

\path[draw=drawColor,line width= 0.6pt,fill=fillColor] (2387.77,475.38) --
	(2387.77,472.57) --
	(2494.27,472.57) --
	(2494.27,475.38) --
	(2387.77,475.38) --
	cycle;

\path[draw=drawColor,line width= 0.6pt,line join=round] (2756.58,477.02) -- (2756.58,480.54);

\path[draw=drawColor,line width= 0.6pt,line join=round] (2756.58,474.55) -- (2756.58,470.94);

\path[draw=drawColor,line width= 0.6pt,fill=fillColor] (2703.33,477.02) --
	(2703.33,474.55) --
	(2809.83,474.55) --
	(2809.83,477.02) --
	(2703.33,477.02) --
	cycle;

\path[draw=drawColor,line width= 0.6pt,line join=round] (3072.15,413.81) -- (3072.15,442.83);

\path[draw=drawColor,line width= 0.6pt,line join=round] (3072.15,387.91) -- (3072.15,352.02);

\path[draw=drawColor,line width= 0.6pt,fill=fillColor] (3018.90,413.81) --
	(3018.90,387.91) --
	(3125.40,387.91) --
	(3125.40,413.81) --
	(3018.90,413.81) --
	cycle;

\path[draw=drawColor,line width= 0.6pt,line join=round] (3387.71,456.26) -- (3387.71,492.47);

\path[draw=drawColor,line width= 0.6pt,line join=round] (3387.71,430.21) -- (3387.71,391.36);

\path[draw=drawColor,line width= 0.6pt,fill=fillColor] (3334.46,456.26) --
	(3334.46,430.21) --
	(3440.96,430.21) --
	(3440.96,456.26) --
	(3334.46,456.26) --
	cycle;

\path[draw=drawColor,line width= 0.6pt,line join=round] (4018.84,564.54) -- (4018.84,598.15);

\path[draw=drawColor,line width= 0.6pt,line join=round] (4018.84,507.96) -- (4018.84,474.79);

\path[draw=drawColor,line width= 0.6pt,fill=fillColor] (3965.59,564.54) --
	(3965.59,507.96) --
	(4072.09,507.96) --
	(4072.09,564.54) --
	(3965.59,564.54) --
	cycle;
\definecolor{drawColor}{RGB}{255,240,245}

\path[draw=drawColor,line width= 0.0pt,line join=round] (2539.63,206.30) -- (2539.63,722.07);

\path[draw=drawColor,line width= 0.0pt,line join=round] (2855.20,206.30) -- (2855.20,722.07);

\path[draw=drawColor,line width= 0.0pt,line join=round] (3170.76,206.30) -- (3170.76,722.07);

\path[draw=drawColor,line width= 0.0pt,line join=round] (3486.33,206.30) -- (3486.33,722.07);

\path[draw=drawColor,line width= 0.0pt,line join=round] (3801.89,206.30) -- (3801.89,722.07);
\end{scope}
\begin{scope}
\path[clip] (150.65,1294.75) rectangle (2107.15,1445.40);
\definecolor{drawColor}{gray}{0.10}

\node[text=drawColor,anchor=base,inner sep=0pt, outer sep=0pt, scale=  8.00] at (1128.90,1342.53) {ADNI};
\end{scope}
\begin{scope}
\path[clip] (2192.51,1294.75) rectangle (4149.01,1445.40);
\definecolor{drawColor}{gray}{0.10}

\node[text=drawColor,anchor=base,inner sep=0pt, outer sep=0pt, scale=  8.00] at (3170.76,1342.53) {OASIS-3};
\end{scope}
\begin{scope}
\path[clip] (  0.00,778.98) rectangle (150.65,1294.75);
\definecolor{drawColor}{gray}{0.10}

\node[text=drawColor,rotate= 90.00,anchor=base,inner sep=0pt, outer sep=0pt, scale=  8.00] at (102.87,1036.86) {FIRST};
\end{scope}
\begin{scope}
\path[clip] (  0.00,206.30) rectangle (150.65,722.07);
\definecolor{drawColor}{gray}{0.10}

\node[text=drawColor,rotate= 90.00,anchor=base,inner sep=0pt, outer sep=0pt, scale=  8.00] at (102.87,464.19) {FASTSURFER};
\end{scope}
\begin{scope}
\path[clip] (  0.00,  0.00) rectangle (4336.20,1445.40);
\definecolor{drawColor}{RGB}{0,0,0}

\node[text=drawColor,anchor=base,inner sep=0pt, outer sep=0pt, scale=  8.00] at (339.99,106.20) {DEFACE};

\node[text=drawColor,anchor=base,inner sep=0pt, outer sep=0pt, scale=  8.00] at (971.12,106.20) {FACE MASK v1};

\node[text=drawColor,anchor=base,inner sep=0pt, outer sep=0pt, scale=  8.00] at (1602.25,106.20) {CP-GAN};
\end{scope}
\begin{scope}
\path[clip] (  0.00,  0.00) rectangle (4336.20,1445.40);
\definecolor{drawColor}{RGB}{0,0,0}

\node[text=drawColor,anchor=base,inner sep=0pt, outer sep=0pt, scale=  8.00] at (655.55, 15.55) {QUICKSHEAR};

\node[text=drawColor,anchor=base,inner sep=0pt, outer sep=0pt, scale=  8.00] at (1286.68, 15.55) {FACE MASK v2};

\node[text=drawColor,anchor=base,inner sep=0pt, outer sep=0pt, scale=  8.00] at (1917.81, 15.55) {CP-MAE};
\end{scope}
\begin{scope}
\path[clip] (  0.00,  0.00) rectangle (4336.20,1445.40);
\definecolor{drawColor}{RGB}{0,0,0}

\node[text=drawColor,anchor=base,inner sep=0pt, outer sep=0pt, scale=  8.00] at (2381.85,106.20) {DEFACE};

\node[text=drawColor,anchor=base,inner sep=0pt, outer sep=0pt, scale=  8.00] at (3012.98,106.20) {FACE MASK v1};

\node[text=drawColor,anchor=base,inner sep=0pt, outer sep=0pt, scale=  8.00] at (3644.11,106.20) {CP-GAN};
\end{scope}
\begin{scope}
\path[clip] (  0.00,  0.00) rectangle (4336.20,1445.40);
\definecolor{drawColor}{RGB}{0,0,0}

\node[text=drawColor,anchor=base,inner sep=0pt, outer sep=0pt, scale=  8.00] at (2697.41, 15.55) {QUICKSHEAR};

\node[text=drawColor,anchor=base,inner sep=0pt, outer sep=0pt, scale=  8.00] at (3328.54, 15.55) {FACE MASK v2};

\node[text=drawColor,anchor=base,inner sep=0pt, outer sep=0pt, scale=  8.00] at (3959.67, 15.55) {CP-MAE};
\end{scope}
\begin{scope}
\path[clip] (  0.00,  0.00) rectangle (4336.20,1445.40);
\definecolor{drawColor}{RGB}{0,0,0}

\node[text=drawColor,anchor=base west,inner sep=0pt, outer sep=0pt, scale=  8.00] at (4194.01,774.87) {0.80};

\node[text=drawColor,anchor=base west,inner sep=0pt, outer sep=0pt, scale=  8.00] at (4194.01,892.09) {0.85};

\node[text=drawColor,anchor=base west,inner sep=0pt, outer sep=0pt, scale=  8.00] at (4194.01,1009.32) {0.90};

\node[text=drawColor,anchor=base west,inner sep=0pt, outer sep=0pt, scale=  8.00] at (4194.01,1126.54) {0.95};

\node[text=drawColor,anchor=base west,inner sep=0pt, outer sep=0pt, scale=  8.00] at (4194.01,1243.76) {1.00};
\end{scope}
\begin{scope}
\path[clip] (  0.00,  0.00) rectangle (4336.20,1445.40);
\definecolor{drawColor}{RGB}{0,0,0}

\node[text=drawColor,anchor=base west,inner sep=0pt, outer sep=0pt, scale=  8.00] at (4194.01,202.19) {0.80};

\node[text=drawColor,anchor=base west,inner sep=0pt, outer sep=0pt, scale=  8.00] at (4194.01,319.42) {0.85};

\node[text=drawColor,anchor=base west,inner sep=0pt, outer sep=0pt, scale=  8.00] at (4194.01,436.64) {0.90};

\node[text=drawColor,anchor=base west,inner sep=0pt, outer sep=0pt, scale=  8.00] at (4194.01,553.86) {0.95};

\node[text=drawColor,anchor=base west,inner sep=0pt, outer sep=0pt, scale=  8.00] at (4194.01,671.08) {1.00};
\end{scope}
\begin{scope}
\path[clip] (  0.00,  0.00) rectangle (4336.20,1445.40);
\definecolor{fillColor}{RGB}{255,255,255}

\path[fill=fillColor] (1829.35,1307.16) rectangle (2470.31,1434.72);
\end{scope}
\begin{scope}
\path[clip] (  0.00,  0.00) rectangle (4336.20,1445.40);
\definecolor{drawColor}{RGB}{102,194,165}

\path[draw=drawColor,line width= 0.6pt] (1925.87,1342.68) --
	(1925.87,1353.28);

\path[draw=drawColor,line width= 0.6pt] (1925.87,1388.60) --
	(1925.87,1399.20);
\definecolor{fillColor}{RGB}{102,194,165}

\path[draw=drawColor,line width= 0.6pt,fill=fillColor] (1912.32,1353.28) rectangle (1939.42,1388.60);

\path[draw=drawColor,line width= 0.6pt] (1912.32,1370.94) --
	(1939.42,1370.94);
\end{scope}
\begin{scope}
\path[clip] (  0.00,  0.00) rectangle (4336.20,1445.40);
\definecolor{drawColor}{RGB}{252,141,98}

\path[draw=drawColor,line width= 0.6pt] (2217.90,1342.68) --
	(2217.90,1353.28);

\path[draw=drawColor,line width= 0.6pt] (2217.90,1388.60) --
	(2217.90,1399.20);
\definecolor{fillColor}{RGB}{252,141,98}

\path[draw=drawColor,line width= 0.6pt,fill=fillColor] (2204.35,1353.28) rectangle (2231.45,1388.60);

\path[draw=drawColor,line width= 0.6pt] (2204.35,1370.94) --
	(2231.45,1370.94);
\end{scope}
\begin{scope}
\path[clip] (  0.00,  0.00) rectangle (4336.20,1445.40);
\definecolor{drawColor}{RGB}{0,0,0}

\node[text=drawColor,anchor=base west,inner sep=0pt, outer sep=0pt, scale=  8.00] at (1993.94,1343.39) {$128^3$};
\end{scope}
\begin{scope}
\path[clip] (  0.00,  0.00) rectangle (4336.20,1445.40);
\definecolor{drawColor}{RGB}{0,0,0}

\node[text=drawColor,anchor=base west,inner sep=0pt, outer sep=0pt, scale=  8.00] at (2285.97,1343.39) {$256^3$};
\end{scope}
\end{tikzpicture}

%% file: 5_conclusion.tex
\section{Conclusion}
In this work, we introduce \model{CP-MAE}, a new MRI de-identification technique that uniquely anonymizes scans by focusing solely on the skull, employing a generative model. Our approach combines dual instances of \model{VQ-VAE} with an \model{MAE} that operates in a highly-compressed space. This setup enables \model{CP-MAE} to stochastically generate high-quality MR scans with the original brain content but significantly altered external appearance. Notably, we enhance the 3D de-identification resolution capability of \model{CP-GAN} from $128^3$ to $256^3$, achieving an eightfold voxel increase. A user study confirmed \model{CP-MAE}'s superior de-identification efficacy over competing methods. Furthermore, our analysis on standard brain analysis tasks reveals that \model{CP-MAE} minimally impacts diagnostic accuracy. Future explorations will extend to the CT imaging domain.